\documentclass{article} 
\usepackage[preprint]{colm2026_conference}

\usepackage{microtype}
\usepackage{hyperref}
\usepackage{url}
\usepackage{booktabs}
\usepackage{graphicx}
\usepackage{placeins}
\usepackage{amsmath} 
\usepackage{enumitem}
\usepackage{adjustbox}
\usepackage{gensymb}

\usepackage{lineno}

\definecolor{darkblue}{rgb}{0, 0, 0.5}
\hypersetup{colorlinks=true, citecolor=darkblue, linkcolor=darkblue, urlcolor=darkblue}

\title{Closing the Confidence-Faithfulness Gap in Large Language Models}

\author{\centerline{Miranda Muqing Miao\thanks{Corresponding author: \texttt{miaom@seas.upenn.edu}. Department of Computer and Information Science.},\quad Lyle Ungar} \\ \\ \centerline{ University of Pennsylvania}}
\begin{document}

\ifcolmsubmission
\linenumbers
\fi

\maketitle

\begin{abstract}
Large language models (LLMs) tend to verbalize confidence scores that are largely detached from their actual accuracy, yet the geometric relationship governing this behavior remain poorly understood. In this work, we present a mechanistic interpretability analysis of verbalized confidence, using linear probes and contrastive activation addition (CAA) steering to show that calibration and verbalized confidence signals are encoded linearly but are orthogonal to one another — a finding consistent across three open-weight models and four datasets. Interestingly, when models are prompted to simultaneously reason through a problem and verbalize a confidence score, the reasoning process disrupts the verbalized confidence direction, exacerbating miscalibration. We term this the "Reasoning Contamination Effect." Leveraging this insight, we introduce a two-stage adaptive steering pipeline that reads the model's internal accuracy estimate and steers verbalized output to match it, substantially improving calibration alignment across all evaluated models.
\end{abstract}




\section{Introduction}

Large language models have shown to be systematically overconfident. This miscalibration primarily takes two forms: at the token level, where output probabilities are poorly calibrated despite high accuracy \citep{guo2017calibration, desai2020calibration}, and at the \textbf{verbalized level}, where models cluster their verbal confidence scores near the top of the range regardless of actual performance \citep{lin2022teaching, kadavath2022language, xiong2024can}. Instruction tuning and RLHF exacerbate the problem, compressing verbalized confidence even further toward high certainty \citep{tian2023just, leng2025taming}. Of these two failure modes, verbalized confidence is particularly consequential for safe deployment. It is the primary natural language channel through which the average user receives uncertainty information. When a model tells a physician ``I am 95\% confident'' about a diagnosis it answers correctly only 40\% of the time, the downstream consequences can be catastrophic.

We argue that verbalized miscalibration is not caused by a lack of internal knowledge but by a failure to read out signals that are already present. The information needed for faithful confidence statements exists in the residual stream; the generation process simply fails to use it. This understanding shifts the question from ``how do we teach models to be calibrated?'' to ``how do we correct the readout?''

A growing body of mechanistic-interpretability research has shown that high-level semantic and behavioral properties are encoded as linear directions in the residual stream. Linear probes recover truth and falsehood from internal activations \citep{burns2023discovering, marks2024geometry, azaria2023internal}, and steering vectors along these directions causally shift model behavior at inference time for truthfulness \citep{li2023iti}, broad behavioral traits \citep{zou2023representation, turner2024activation, panickssery2024steering}, and refusal \citep{arditi2024refusal}. Simultaneous work has begun extending this lens to verbalized confidence. \citet{kumaran2026compute} show that verbal confidence is cached at answer-adjacent positions and retrieved later. \citet{advice2026} identify ``answer-independence'' as a driver of overconfidence and propose a fine-tuning fix. These studies establish that verbalized confidence has a nontrivial internal presence, yet a core question remains unanswered: what is the geometric relationship between the model's internal accuracy signal and its verbalized confidence, and can that relationship be leveraged to improve calibration? 

Existing methods for improving verbalized-confidence calibration treat the model as a black box. Prompt-engineering strategies elicit better-calibrated scores by asking models to consider top-$K$ alternatives \citep{tian2023just} or by aggregating across multiple response samples \citep{xiong2024can}. The most closely related prompting work, SteerConf \citep{zhou2025steerconf}, shifts verbalized confidence through a range of cautious-to-confident prompt framings and aggregates the resulting scores. Training-based approaches fine-tune models to express calibrated scores using proper scoring rules \citep{li2025conftuner} or RL reward shaping 
\citep{baniharouni2026rewarding}. All of these methods manipulate the input or retrain the model without leveraging existing signals at the representational level. By contrast, our two-stage pipeline reads the model's internal accuracy estimate and steers the output to match it, achieving substantially lower calibration error than both unsteered verbalized confidence and SteerConf across all evaluated models.

Our main contributions are:
\begin{itemize}

\item \textbf{Geometric dissociation.} Models encode well-calibrated accuracy information in a linearly accessible direction, but verbalized confidence occupies a separate, nearly orthogonal direction (cosine similarity $<0.04$). The model ``knows'' when it is likely wrong, but the generation process fails to surface this signal.

\item \textbf{Reasoning contamination.} When the model solves a problem and rates its confidence jointly, the confidence and accuracy directions shift from weakly aligned to sharply opposed (cosine similarity dropping from $+0.26$ to $-0.63$), meaning joint prompting actively inverts the relationship between what the model knows and what it says.

\item \textbf{Steering-based calibration.} Contrastive activation addition produces causally controlled shifts in verbalized confidence that generalize across datasets and transfer from base to instruction-tuned models. We introduce a two-stage adaptive steering pipeline that reads the model's internal accuracy estimate and steers verbalized output to match, improving calibration by $4$--$7\times$.

\end{itemize}

\section{Method}
\label{sec:method}
This section describes the two methodological tools that underpin our analysis. We first introduce gold calibration linear probing, which tests whether accuracy information is linearly accessible in the residual stream. We then describe contrastive activation steering, which constructs steering vectors that causally shift verbalized confidence at inference time. The remaining paragraphs detail the datasets, models, activation extraction procedure, and prompt design used throughout.

\subsection{Gold Calibration Linear Probing}
\label{sec:probing}

To test whether calibration information is linearly accessible in model activations, we train ridge regression probes on extracted activation vectors. For \emph{gold calibration probing}, we use activations from the \textbf{pure correctness} prompt and regress against binary correctness labels or binned empirical accuracy (the fraction of times the model answers a question correctly across 50 samples with different random seeds). We sweep over a broad range of $\ell_2$ regularization strengths and select the value that maximizes \textit{validation} performance. 

\subsection{Contrastive Activation Steering}
\label{sec:steering}

To move beyond correlation and establish a causal link between activation directions and verbalized confidence, we apply contrastive activation addition (CAA)~\citep{turner2023activation}. We elicit the same set of questions under $K=11$ prompt framings that span a wide range of instructed confidence levels, collecting hidden-state activations $\mathbf{h}_{q,k}^{(\ell)} \in \mathbb{R}^d$ at layer $\ell$ for question $q$ under framing $k$. Each instance is paired with its parsed verbalized confidence $c_{q,k} \in [0,1]$. The exact $K$ prompts used are shown in the Appendix. 

We partition instances into a \emph{high-confidence} set $\mathcal{H}_q = \{k : c_{q,k} > \tau_{\mathrm{hi}}\}$ and a \emph{low-confidence} set $\mathcal{L}_q = \{k : c_{q,k} < \tau_{\mathrm{lo}}\}$, with $\tau_{\mathrm{hi}} = 0.75$ and $\tau_{\mathrm{lo}} = 0.25$. For each question $q$ that contains at least one instance in both sets, we compute a per-question contrast:
\begin{equation}
  \boldsymbol{\delta}_q^{(\ell)}
  = \frac{1}{|\mathcal{H}_q|}\sum_{k \in \mathcal{H}_q} \mathbf{h}_{q,k}^{(\ell)}
  - \frac{1}{|\mathcal{L}_q|}\sum_{k \in \mathcal{L}_q} \mathbf{h}_{q,k}^{(\ell)}.
\end{equation}
The steering vector is then obtained by averaging over all qualifying questions $\mathcal{Q}$:
\begin{equation}
  \mathbf{v}^{(\ell)} = \frac{1}{|\mathcal{Q}|}\sum_{q \in \mathcal{Q}} \boldsymbol{\delta}_q^{(\ell)}.
\end{equation}
Because each $\boldsymbol{\delta}_q$ is computed \emph{within} a single question, this design controls for confounds such as question difficulty, topic, and prompt framing, isolating the component of the activation that varies specifically with expressed confidence.

At inference time, we inject the steering vector into the residual stream during autoregressive generation. Let $\mathbf{h}_t^{(\ell)}$ denote the hidden state at layer $\ell$ and generation step $t$. The steered activation is:
\begin{equation}
  \tilde{\mathbf{h}}_t^{(\ell)} = \mathbf{h}_t^{(\ell)} + \alpha \, \hat{\mathbf{v}}^{(\ell)},
\end{equation}
where $\hat{\mathbf{v}}^{(\ell)} = \frac{\mathbf{v}^{(\ell)}}{\|\mathbf{v}^{(\ell)}\|} \cdot \bar{n}^{(\ell)}$ is the steering vector normalized to unit length and rescaled by the mean activation norm $\bar{n}^{(\ell)}$ at layer $\ell$, and $\alpha \in \mathbb{R}$ controls steering strength. We evaluate three injection sites: the last prompt token only, every answer token, and both jointly. Steering at the answer-token position yields the most stable results, slightly outperforming the combined condition; we therefore report answer-token steering throughout. The steering layer, variant, and strength are selected on a validation split, and all steered generations use temperature $T = 1.0$, matching the activation-collection setting. 

\paragraph{Datasets:} We evaluate on four question-answering benchmarks that span mathematical reasoning, broad knowledge, and truthfulness: MATH \citep{hendrycks2021math}, MMLU \citep{hendrycks2021mmlu}, TriviaQA \citep{joshi2017triviaqa}, and TruthfulQA \citep{lin2022truthfulqa}. Each dataset contains three non-overlapping splits: a training split for extracting activations and fitting probes, a validation split for selecting optimal steering layers and strengths, and a held-out test split for final evaluation. 

\paragraph{Models:} We conduct experiments across three model families: Llama-3.1-8B \citep{touvron2023llama}, Qwen2.5-7B \citep{qwen2025qwen25}, and Mistral-7B-v0.1 \citep{jiang2023mistral}. For each family, we analyze both the base (pretrained) model and its corresponding instruction-tuned (instruct) variant. 

\paragraph{Activation Extraction:} We extract residual stream activations after the MLP sublayer at each transformer layer. For each input, we record the hidden state at two positions: the final prompt token (\emph{prompt completion}) and the final generated token (\emph{answer completion}). Both extraction points yield similar steering vectors and downstream effects; we use prompt-completion activations throughout, as they can be obtained before generation begins and are therefore more practical for inference-time interventions. All generations use sampling temperature $T{=}1.0$ to elicit the model's default output distribution.

\paragraph{Prompt Design:} We utilize three prompt types to disentangle the model's representations of answer correctness and expressed confidence. The \textbf{pure correctness} prompt asks the model only to answer the question, with no mention of confidence. The \textbf{pure confidence} prompt asks the model only to state how confident it is in answering a given question correctly, without producing the answer. The \textbf{joint} prompt asks the model to both express its confidence and provide an answer. This design is \textbf{critical} for analyzing the relationship between accuracy and verbalized confidence in Sec.~\ref{sec:reasoning_contamination} of the paper. Exactly prompts are shown in the Appendix.

\section{Results}
\label{sec:results}

We organize our results around three questions. First, are accuracy and verbalized confidence linearly encoded, and how do they relate geometrically? We show both signals are linearly decodable but nearly orthogonal, and that joint prompting inverts their relationship (\S\ref{sec:gold_calibration}--\ref{sec:reasoning_contamination}). Second, is the verbalized confidence direction causally active and general? We show that steering vectors shift verbalized confidence in a controlled manner, generalize across datasets, and transfer from base to instruction-tuned models (\S\ref{sec:monotonic}--\ref{sec:cross_model}). Third, can we close the calibration gap? We introduce an adaptive steering pipeline that meaningfully improves ECE, brier score, and MAE. (\S\ref{sec:two_stage_steering}).

\subsection{Gold Calibration Information Is Linearly Encoded}
\label{sec:gold_calibration}

\begin{figure}[!htbp]
    \centering
    \includegraphics[width=0.58\columnwidth]{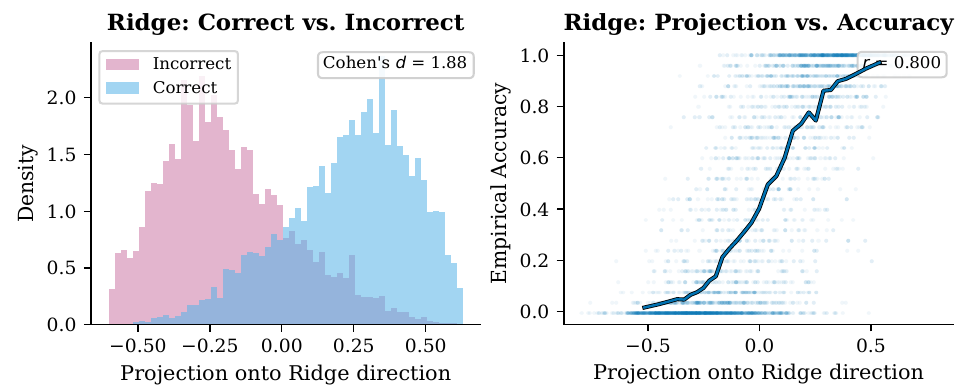}
\caption{\textbf{Ridge probe projection at layer 21 (Qwen-2.5-7B-Base).}
\textit{Left:} Distribution of activations projected onto the probe weight vector,
separated by correct (blue) and incorrect (pink) answers (Cohen's $d = 1.88$).
\textit{Right:} The same scalar projection plotted against binned empirical accuracy
($r = 0.80$).
\textbf{Takeaway:} The model encodes well-calibrated accuracy information in a single
linear direction, even when never asked about confidence.}
    \label{fig:gold_calibration}
\end{figure}

We extract activations under a \textbf{pure correctness} prompt, one that asks the model to produce only an answer, with no mention of confidence, then train a ridge regression probe to predict empirical accuracy: the fraction of times the model answers a given question correctly across repeated samples. As shown in Figure~\ref{fig:gold_calibration}, a single linear direction in the residual stream cleanly separates correct from incorrect responses (Cohen's $d = 1.88$) and, more importantly, tracks graded empirical accuracy at $r = 0.80$. The model thus encodes well-calibrated uncertainty information in a linearly accessible direction, even when it is never prompted to express confidence; the calibration signal is present in the activations, but the generation process fails to surface it.

\subsection{Verbalized Confidence Is Linearly Separable}
\label{sec:verbal_linear}
\begin{figure}[t]
    \centering
    \includegraphics[width=0.93\columnwidth]{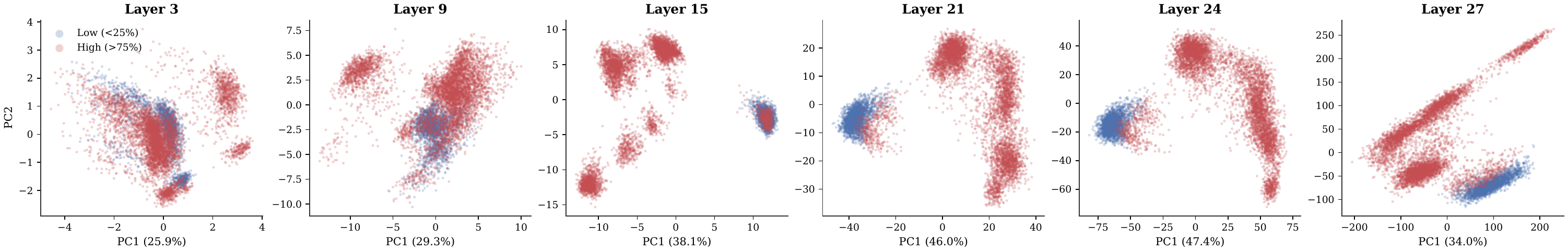}
\caption{\textbf{High and low verbalized confidence occupy distinct regions of
activation space (25th vs.\ 75th percentile split).} First principal component of
activations from the \textbf{pure confidence} prompt, colored by whether the model
verbalized high or low confidence.
\textbf{Takeaway:} Verbalized confidence is linearly separable in later layers,
confirming that the model constructs a dedicated confidence representation during
processing.}

    \label{fig:pca_25_75}
\end{figure}

Using activations from the \textbf{pure confidence} prompt, we project onto the first principal component and color each point by whether the model expressed high or low confidence. Figure~\ref{fig:pca_25_75} reveals clear linear separability between high- and low-confidence activations in later layers, suggesting that the model progressively constructs a linearly separable representation of its own confidence. To quantify this effect, we train linear ridge regression and report train and test time $R^2$ in Figure~\ref{fig:direction_analysis} a and b. The natural next question is whether they share the same direction or dissociated, which would explain verbalized miscalibration. 

\subsection{Verbalized Confidence and Accuracy Occupy Orthogonal Directions}
\label{sec:direction_analysis}

\begin{figure}[!htbp]
    \centering
    \includegraphics[width=0.70\columnwidth]{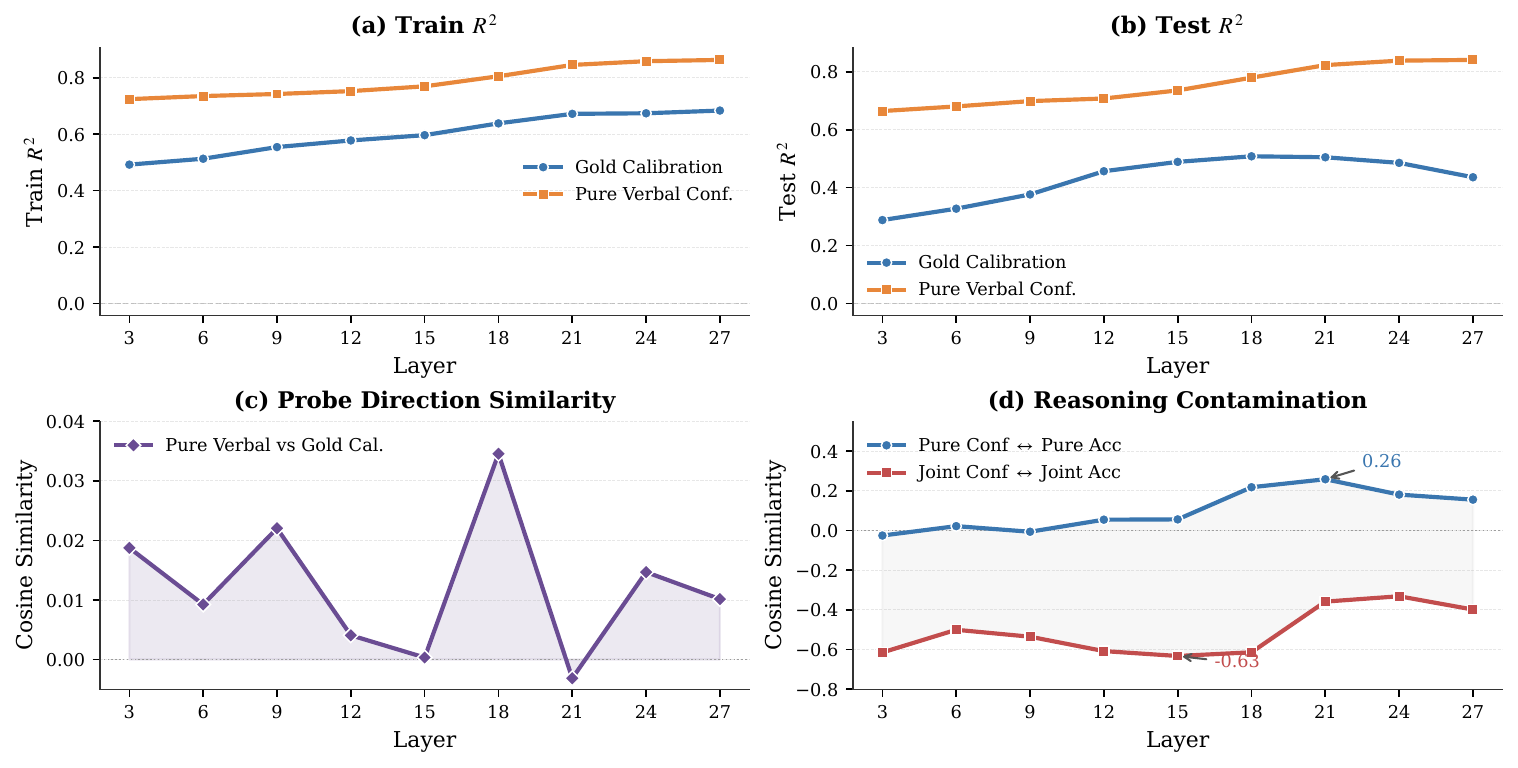}
\caption{\textbf{Probe fit and directional analysis across layers (Qwen-2.5-7B-Base).}
\textbf{(a, b)} Train and test $R^2$ of ridge probes predicting empirical accuracy
(gold calibration, blue) and verbalized confidence (pure verbal, orange).
\textbf{(c)} Cosine similarity between the two probe weight vectors (pure verbal
vs.\ gold calibration).
\textbf{(d)} Cosine similarity between contrastive confidence and accuracy directions,
computed separately under the pure confidence prompt (blue) and the joint
solve-and-rate prompt (red). Shaded region indicates the gap between the two
conditions, the \textbf{reasoning contamination effect}.
\textbf{Takeaway:} Accuracy and confidence are encoded in nearly orthogonal directions
(cosine similarity $< 0.04$), and joint prompting inverts their relationship
(from $+0.26$ to $-0.63$).}
    \label{fig:direction_analysis}
\end{figure}

Although both gold calibration and pure verbalized confidence are both individually predictable using linear probes, with test $R^2$ reaching 0.55 and 0.85 respectively (Figure~\ref{fig:direction_analysis}a, b), the directions that encode these two signals are nearly orthogonal. Figure~\ref{fig:direction_analysis}(c) shows that the cosine similarity between the two ridge probe weight vectors remains below 0.04 across all layers. The model thus likely maintains separate linear subspaces for ``how likely am I to be correct'' and ``how confident do I say I am.'' This dissociation is consistent with our observation that base models verbalize poorly calibrated confidence despite encoding well-calibrated accuracy information internally (\S\ref{sec:gold_calibration}). To further illustrate the clear orthogonality phenomenon in higher dimensions, we include four distinct subspace-level analyses in Appendix~\ref{sec:subspace_cca}. 

\subsection{Reasoning Contamination Inverts the Verbalized Confidence--Accuracy Relationship}
\label{sec:reasoning_contamination}
Does the relationship between the confidence and accuracy directions depend on how the model is prompted? We define two setups: a \emph{pure confidence} condition, where the model rates its confidence on answering a question correctly without solving the problem, and a \emph{joint} condition, where the model solves the problem and rates its confidence in the same generation. For each condition and layer, we extract a contrastive confidence direction (mean activation of high-confidence instances minus mean of low-confidence instances) and a contrastive accuracy direction (mean of high-accuracy instances minus mean of low-accuracy instances), then measure the cosine similarity between them.

Figure~\ref{fig:direction_analysis}(d) demonstrate the layer-wise output. Under the pure condition (blue), the confidence and accuracy directions start out as completely orthogonal and become are weakly positive, reaching $+0.26$ at layer 21. This indicates that when the model assesses confidence in isolation, its confidence representation partially aligns with genuine competence. Under the joint condition (red), the relationship inverts. The two directions are completely anti-correlated across all layers, reaching $-0.63$ at layer 15. When the model reasons about a problem and rates its confidence simultaneously, the direction encoding verbalized confidence actively opposes the direction encoding correctness. We coin this the \emph{reasoning contamination effect}. Joint prompts produce representations in which confidence and accuracy point in opposite directions.

This effect is prominent during CAA steering. When we apply verbalized confidence steering using \textbf{joint} solve-and-rate generation, the more we influenced the verbalized confidence output, the more we erode the model's accuracy performance. This motivates the design of our two-stage pipeline in Sec.~\ref{sec:two_stage_steering}, where we improve verbalized confidence calibration using \textbf{pure confidence} prompts, leaving the model's problem-solving pass entirely unperturbed in a separate run. 

\subsection{Steering Produces Principled Shifts in Verbalized Confidence}
\label{sec:monotonic}

The preceding sections establish the direction of gold calibration and verbalized confidence. But is the verbalized confidence direction merely a statistical pattern, or is it causally active? We apply CAA steering vectors constructed from top-versus-bottom quartile activations under the \textbf{pure confidence} prompt. Steering is applied during generation under the same prompt condition, so that the model is only verbalizing confidence, not solving the problem. Table~\ref{tab:monotonic} presents the central causal result: scaling the steering vector produces a clean positive shift in verbalized confidence across MATH, TriviaQA, and TruthfulQA. Positive scaling increases verbalized confidence, negative scaling decreases it, and the relationship is approximately linear over a wide range of steering strengths.

\begin{table}[!htbp]
    \centering
\caption{\textbf{Activation steering produces principled shifts in verbalized
confidence.} Mean verbalized confidence as a function of steering strength
(multiples of the CAA vector), evaluated on MATH, TriviaQA, TruthfulQA, and MMLU.
\textbf{Takeaway:} The verbalized confidence direction is causally active, with
positive and negative scaling producing controlled, approximately linear shifts
across all models and datasets. Results increase (→) across columns.}

    \vspace{5pt}
    \label{tab:monotonic}
    \small
    \setlength{\tabcolsep}{4pt}
    \begin{adjustbox}{max width=0.68\textwidth}
    \begin{tabular}{llccccccc}
        \toprule
        & & \multicolumn{3}{c}{\textbf{Neg.\ Steering}} & & \multicolumn{3}{c}{\textbf{Pos.\ Steering}} \\
        \cmidrule(lr){3-5} \cmidrule(lr){7-9}
        \textbf{Model} & \textbf{Dataset} &$\alpha{=}{-}0.75$ & ${-}0.50$ & ${-}0.25$ & $0$ & $0.25$ & $0.50$ & $0.75$ \\
        \midrule
        Mistral-7B      & MATH        & 23.1 & 51.4 & 57.2 & 64.7 & 64.7 & 88.6 & 99.0 \\
        (layer 27)      & TriviaQA    & 34.5 & 38.4 & 42.8 & 56.2 & 67.1 & 83.1 & 94.5 \\
                         & TruthfulQA  & 24.0 & 35.3 & 41.5 & 54.2 & 77.5 & 82.2 & 92.6 \\
                         & MMLU        & 25.8 & 35.5 & 45.6 & 60.0 & 60.8 & 88.0 & 97.9 \\
        \midrule
        Llama-3.1-8B    & MATH        & 18.2 & 29.3 & 37.4 & 45.3 & 55.1 & 71.4 & 85.8 \\
        (layer 24)      & TriviaQA    & 17.8 & 26.8 & 30.6 & 48.1 & 56.5 & 71.2 & 87.7 \\
                         & TruthfulQA  & 38.7 & 40.8 & 49.3 & 51.8 & 56.8 & 64.9 & 85.9 \\
                         & MMLU        & 18.9 & 21.4 & 28.3 & 48.5 & 65.6 & 73.0 & 88.3 \\
        \midrule
        Qwen2.5-7B\footnotemark  & MATH        & 29.3 & 35.7 & 48.6 & 66.0 & 82.1 & 84.3 & 92.5 \\
        (layer 21)      & TriviaQA    & 45.5 & 53.0 & 55.5 & 60.6 & 63.5 & 65.0 & 67.8 \\
                         & TruthfulQA  & 52.6 & 54.3 & 56.3 & 60.2 & 61.4 & 62.8 & 64.1 \\
                         & MMLU        & 62.2 & 64.1 & 66.3 & 66.6 & 71.3 & 71.1 & 77.0 \\
        \bottomrule
    \end{tabular}
    \end{adjustbox}
    \vspace{-6pt}
\end{table}

\footnotetext{Qwen2.5-7B verbalizes meaningfully higher baseline confidence (~15-20\% higher) than Llama and Mistral during the activation collection process, leaving a narrower dynamic range and less headroom for upward steering.}


\subsection{Cross-Dataset Generalization}
\label{sec:cross_dataset}

A steering vector is most useful if it generalizes beyond the distribution on which it was constructed. We calculate CAA vectors exclusively on MATH activations and evaluate their steering effect on MMLU, TriviaQA, and TruthfulQA without any adaptation. Table~\ref{tab:transfer_dataset} reports the results.

\begin{table}[t]
    \centering
\caption{\textbf{MATH-derived steering vectors transfer across datasets.} Mean
verbalized confidence (\%) under varying steering magnitudes, using a CAA vector
trained only on MATH at layer 21 of Qwen2.5-7B, layer 24 of Llama-3.1-8B, and
layer 27 of Mistral-7B-v0.1.
\textbf{Takeaway:} The confidence direction is domain-general, not an artifact of
mathematical notation or problem format. Results increase (→) across columns.}
    \label{tab:transfer_dataset}
    \small
    \vspace{5pt}
    \begin{adjustbox}{max width=0.68\textwidth}
    \begin{tabular}{lccccccc}
        \toprule
        & \multicolumn{3}{c}{\textbf{Negative Steering}} & & \multicolumn{3}{c}{\textbf{Positive Steering}} \\
        \cmidrule(lr){2-4} \cmidrule(lr){6-8}
        \textbf{Target Dataset} & $\alpha{=}{-}0.75$ & ${-}0.50$ & ${-}0.25$ & $0$ & $0.25$ & $0.50$ & $0.75$ \\
        \midrule
        \multicolumn{8}{l}{\textbf{\textit{Llama-3.1-8B}}} \\
        \quad MMLU             & 3.6 & 6.4 & 13.8 & 18.7 & 34.3 & 56.6 & 76.2 \\
        \quad TriviaQA         & 6.0 & 7.2 & 15.6 & 24.2 & 30.4 & 57.5 & 76.5 \\
        \quad TruthfulQA       & 5.6 & 9.7 & 19.5 & 21.9 & 35.7 & 57.7 & 75.7 \\
        \midrule
        \multicolumn{8}{l}{\textbf{\textit{Mistral-7B-v0.1}}} \\
        \quad MMLU             & 8.1 & 18.5 & 22.6 & 33.8 & 50.5 & 74.7 & 86.4 \\
        \quad TriviaQA         & 10.8 & 18.5 & 26.1 & 41.0 & 54.6 & 75.3 & 85.0 \\
        \quad TruthfulQA       & 12.9 & 20.9 & 27.5 & 43.2 & 58.6 & 73.5 & 85.4 \\
        \midrule
        \multicolumn{8}{l}{\textbf{\textit{Qwen2.5-7B}}} \\
        \quad MMLU             & 58.3 & 59.1 & 64.2 & 64.0 & 67.1 & 66.3 & 68.8 \\
        \quad TriviaQA         & 52.7 & 54.9 & 57.9 & 59.2 & 60.6 & 61.8 & 64.3 \\
        \quad TruthfulQA       & 52.2 & 54.3 & 59.1 & 59.6 & 59.9 & 61.6 & 62.9 \\
        \bottomrule
    \end{tabular}
    \end{adjustbox}
    \vspace{-5pt}
\end{table}

The MATH-derived vector produces consistent directional shifts across all target datasets. This cross-dataset generalization indicates that the confidence direction is not an artifact of MATH-specific features such as mathematical notation or problem format. Instead, it reflects a shared, domain-general mechanism through which language models represent and express confidence.

\subsection{Base-to-Instruct Transfer}
\label{sec:cross_model}

Finally, we test whether confidence directions extracted from base models can steer the verbalized confidence of their instruction-tuned counterparts. This experiment is motivated by the observation that instruct models exhibit more severe overconfidence than base models, suggesting that post-training procedures may suppress or distort the confidence signal that is present in the base model.

\begin{table}[t]
    \centering
\caption{\textbf{Base model steering vectors modulate instruct model confidence.}
Steering vectors extracted from base models applied to their corresponding instruct
variants. 
\textbf{Takeaway:} The confidence direction partially survives post-training,
suggesting that instruct-model overconfidence reflects a readout failure rather than
loss of the underlying signal. Results increase (→) across columns.}
    \label{tab:transfer_base_instruct}
    \small
    \setlength{\tabcolsep}{11pt}
    \vspace{5pt}
    \begin{adjustbox}{max width=0.94\textwidth}
    \begin{tabular}{lccccccc}
        \toprule
        & \multicolumn{3}{c}{\textbf{Neg.\ Steering}} & & \multicolumn{3}{c}{\textbf{Pos.\ Steering}} \\
        \cmidrule(lr){2-4} \cmidrule(lr){6-8}
        \textbf{Model} & $\alpha{=}{-}0.75$ & ${-}0.50$ & ${-}0.25$ & $0$ & $0.25$ & $0.50$ & $0.75$ \\
        \midrule
        Qwen2.5-7B-Inst.       & 38.6 & 75.5 & 86.5 & 88.4 & 90.3 & 94.1 & 98.6 \\
        Llama-3.1-8B-Inst.     & 11.4 & 39.2 & 85.8 & 87.8 & 92.3 & 94.4 & 95.8 \\
        Mistral-7B-Inst.-v0.1  & 37.6 & 46.7 & 50.0 & 63.2 & 80.5 & 86.6 & 90.2 \\
        \bottomrule
    \end{tabular}
    \end{adjustbox}
    \vspace{-5pt}
\end{table}

Table~\ref{tab:transfer_base_instruct} shows that base-model-derived steering vectors successfully modulate instruct model confidence across all three model families. This result has two implications. First, the linear confidence direction identified in base models is not completely eliminated by post-training; it persists in the instruct model's residual stream in a geometrically compatible form and remains stronger in some models than others. Second, it is possible that the overconfidence exhibited by instruct models is not a consequence of losing the confidence signal entirely, but rather of the generation process failing to read it out faithfully. Thus, steering could provide a direct mechanism to restore confidence control in instruct-tuned models.

\subsection{Adaptive Two-Stage Steering for Verbalized Calibration Improvement}
\label{sec:two_stage_steering}

The previous findings show that activation steering can reliably shift verbalized confidence up or down. We now ask whether it can be used to \emph{improve verbalized calibration}, that is, to make verbalized confidence match empirical accuracy. The challenge is that a single global steering strength cannot calibrate all questions. We address this with a two-stage pipeline that assigns a \emph{per-question} steering strength.

\paragraph{Stage 1: Probe-based target estimation.}
We have demonstrated in \ref{sec:gold_calibration} that our gold calibration probe can effectively predict the empirical accuracy of a question using the internal states of the model at prompt completion (before model starts answering) only. The probe's prediction serves as the target confidence for that question: what the model \emph{should} say, given what its activations reveal about its likelihood of being correct. We apply isotonic regression on a held-out validation set to calibrate the probe outputs.

\paragraph{Stage 2: Adaptive steering.}
\begin{table}[!t]
    \centering
\caption{\textbf{Activation steering improves calibration across all models.}
Expected Calibration Error (ECE), Brier Score, and Mean Absolute Error (MAE) for
four confidence sources on MATH. Bolded numbers indicate the best performing
outcomes.
\textbf{Takeaway:} Adaptive steering effectively reduces ECE relative to
unsteered verbalized confidence and substantially outperforms SteerConf, confirming
that reading the model's internal accuracy signal and steering output to match it
closes much of the faithfulness gap.}
    \vspace{5pt}
    \label{tab:two_stage_calibration_metrics}
    \small
    \setlength{\tabcolsep}{5pt}

\begin{adjustbox}{max width=0.68\textwidth}
     \begin{tabular}{llccc}
        \toprule
        \textbf{Model} & \textbf{Confidence Source} & \textbf{ECE $\downarrow$} & \textbf{Brier $\downarrow$} & \textbf{MAE $\downarrow$} \\
        \midrule
        Llama-3.1-8B    & Logit baseline              & 68.4 & 50.5 & 68.5 \\
                               & Verbalized                  & 14.9 & 8.4 & 20.2 \\
                              & SteerConf                    & 46.3 & 40.6 & 53.6 \\
                               & Verbalized (steered)        & \textbf{3.7} & \textbf{2.4} & \textbf{10.5} \\
                            
        \midrule
        Mistral-7B--v0.1 & Logit baseline              & 73.5 & 57.5 & 73.5 \\
                               & Verbalized                  & 35.1 & 15.9 & 40.0 \\
                               & SteerConf                  & 24.3	& 20.1 &29.1 \\
                               & Verbalized (steered)        & \textbf{3.3} & \textbf{2.1} & \textbf{9.3} \\
        \midrule
        Qwen2.5-7B      & Logit baseline              & 75.0 & 58.3 & 74.7 \\
                               & Verbalized                  & 36.0 & 16.7 & 36.9 \\
                                & SteerConf                   & 19.5 & 25.9 & 41.4 \\
                               & Verbalized (steered)        & \textbf{10.7} & \textbf{3.5} & \textbf{11.8} \\
        \bottomrule
    \end{tabular}

\end{adjustbox}
\vspace{-10pt}
\end{table}


We exclusively apply steering during generation under the \emph{pure confidence} prompt. We sweep steering strength $\alpha \in [-2.0, +2.0]$ with 0.1 increments on validation questions to build a transfer function mapping $\alpha$ to mean verbalized confidence. We invert this function via monotone Piecewise Cubic Hermite Interpolating Polynomial (PCHIP) interpolation: given a target confidence $c^*_q$ for question $q$, the inverse yields the steering strength $\alpha^*_q$ that would, on average, produce that confidence level. Each test question thus receives a \emph{question-specific} $\alpha^*_q$, steering overconfident questions downward and underconfident questions upward. We generate 50 samples per question under adaptive steering and report the mean verbalized confidence as the final estimate. We also generate 50 samples of solutions per question in a separate pass to calculate empirical accuracy and match question-level confidence and accuracy to calculate calibration metrics. 

Table~\ref{tab:two_stage_calibration_metrics} reports calibration metrics for three confidence sources. The logit baseline, the token probability assigned to the predicted answer, is severely miscalibrated across all models (ECE $\geq 68$). Unsteered verbalized confidence improves over logit baseline, but remains far from calibrated. Adaptive steering reduces ECE by 4--7$\times$ relative to unsteered verbalized confidence. Mistral shows the largest improvement, dropping from 35.1 to 3.3 ECE and from 15.9 to 2.1 Brier score. The pattern is consistent across all three metrics: by reading the model's internal estimate of its own competence and steering its verbalized output to match, we are able to close much of the faithfulness gap of 
verbalized confidence .

\section{Related Work}

\paragraph{Verbalized confidence calibration.}
\citet{lin2022teaching} introduced verbalized confidence elicitation, and subsequent work has consistently found that LLMs are systematically overconfident across models, domains, and elicitation strategies \citep{kadavath2022language, xiong2024can, groot2024overconfidence}. Prompting-based remedies attempt to shift this distribution: \citet{tian2023just} ask the model to consider top-$K$ alternatives before scoring, \citet{xiong2024can} aggregate confidence across multiple response samples, and \citet{zhou2025steerconf} interpolate between cautious and confident prompt framings. Training-based approaches take a different route, fine-tuning models to produce calibrated scores via proper scoring rules \citep{li2025conftuner} or RL reward shaping \citep{baniharouni2026rewarding}. Our differs by operating on the representations directly, reading the model's internal accuracy signal and steering the output to match.

\paragraph{Internal representations of confidence.}
A growing body of work shows that LLMs encode uncertainty-relevant information in their hidden states. \citet{burns2023discovering} and \citet{marks2024geometry} recover truth and falsehood via linear probes, \citet{azaria2023internal} detect when models produce false statements from hidden-state classifiers, and \citet{stolfo2024confidence} identify dedicated neurons that regulate token-level output entropy. Concurrent work has begun applying similar tools to verbalized confidence specifically. \citet{kumaran2026compute} use activation patching and steering to show that verbal confidence is cached at answer-adjacent positions and reflects richer signals than token log-probabilities. \citet{advice2026} identify answer-independence as a driver of overconfidence through attention and gradient attribution analysis. Our work differs from these studies in both question and method. Where prior analyses ask \emph{what} verbalized confidence represents or \emph{when} it is computed, we ask \emph{why} it diverges from accuracy. 

\section{Discussion, Limitations, and Conclusion}
\paragraph{Discussion:} We hypothesize that reasoning contamination reflects a conflict between two computationally distinct tasks. Problem-solving is heavily optimized during training, while confidence assessment requires self-evaluation the model has far less practice performing. Under joint prompting, high-magnitude activations along effort-encoding directions appear to be interpreted as engagement rather than difficulty, inflating confidence on precisely the questions the model struggles with most. This explains why separating the two tasks into distinct passes, as our pipeline does, prevents the interference.

\paragraph{Limitation:} Our experiments use 7--8B parameter models, and whether the linear encoding and orthogonality findings hold at larger scales, where representations may occupy higher-dimensional subspaces, remains open. Our evaluation is restricted to question-answering tasks with verifiable answers, extending to open-ended generation would require rethinking how the probe target is constructed. Finally, our pipeline requires a separate generation pass for confidence assessment. Learning to steer within a single forward pass is a natural next step toward practical deployment.

\paragraph{Conclusion:} The central message of this work is that verbalized miscalibration in LLMs is a readout failure, not a knowledge deficit. Models encode gold calibration along a linear direction and verbalized confidence along a separate, nearly orthogonal linear direction. The signal needed to produce faithful confidence statements is present in the residual stream, but the generation process fails to use it. Our two-stage pipeline turns this understanding into a practical intervention: a linear probe reads the model's internal accuracy estimate, and contrastive activation addition steers verbalized output to match, substantially reducing calibration error. The verbalized confidence steering vectors generalize across datasets and transfer from base to instruction-tuned models, confirming that the confidence direction is a stable, general-purpose feature of language model representations. More broadly, this work illustrates a pattern we believe will recur: when a model's outputs are misaligned with its internal representations, the most direct remedy is not retraining or prompt engineering but identifying the internal signal and correcting the readout. Verbalized confidence is one instance of this pattern, and we suspect it is not the last.

\bibliography{colm2026_conference}
\bibliographystyle{colm2026_conference}

\newpage

\appendix

\section{Prompt Design}
\begin{figure*}[!htbp]
\centering
\small
\setlength{\fboxsep}{10pt}

\begin{minipage}[t]{0.32\textwidth}
\centering
\textbf{(a) Pure Correctness Prompt}\\[6pt]
\fbox{\begin{minipage}{\dimexpr\linewidth-2\fboxsep-2\fboxrule}
\raggedright
Solve the following math problem step by step.\\[6pt]
Problem: \{problem\}\\[6pt]
Show your work, then write your final answer on a new line in the format:\\
Answer: [your answer]
\end{minipage}}
\end{minipage}%
\hfill
\begin{minipage}[t]{0.32\textwidth}
\centering
\textbf{(b) Pure Confidence Prompt}\\[6pt]
\fbox{\begin{minipage}{\dimexpr\linewidth-2\fboxsep-2\fboxrule}
\raggedright
Read the following math problem and rate your confidence that you can solve it correctly. Do not solve the problem.\\[6pt]
Problem: \{problem\}\\[6pt]
Rate how confident you are that you can solve this problem correctly on a scale from 0 to 100, where 0 means certainly incorrect and 100 means certainly correct.\\
Confidence:
\end{minipage}}
\end{minipage}%
\hfill
\begin{minipage}[t]{0.32\textwidth}
\centering
\textbf{(c) Joint Prompt}\\[6pt]
\fbox{\begin{minipage}{\dimexpr\linewidth-2\fboxsep-2\fboxrule}
\raggedright
Read the following math problem. First rate your confidence that you can solve it correctly, then solve it step by step.\\[6pt]
Problem: \{problem\}\\[6pt]
Rate how confident you are that you can solve this problem correctly on a scale from 0 to 100, where 0 means certainly incorrect and 100 means certainly correct.\\
Confidence: [0--100]\\[4pt]
Show your work, then write your final answer on a new line in the format:\\
Answer: [your answer]
\end{minipage}}
\end{minipage}
\caption{\textbf{Prompt templates for three elicitation conditions.} (a)~The pure correctness prompt asks the model only to solve the problem, with no mention of confidence. (b)~The pure confidence prompt asks the model only to rate its confidence, without producing a solution. (c)~The joint prompt asks the model to first rate its confidence and then solve the problem. Separating these conditions allows us to isolate the model's confidence representation from the computational process of problem-solving.}
\label{fig:prompts}
\end{figure*}

Fig~\ref{fig:prompts} shows the three types of base prompts used for activations extraction under the three conditions: pure answer elicitation, pure confidence elicitation, and both answer and confidence elicitation.

\begin{table*}[t]
\centering
\caption{Prompt framings used to elicit diverse verbalized confidence levels ($K{=}11$). Each framing appends a ``Note'' to the base prompt (see below). The \textsc{Vanilla} framing appends no note. Target ranges are approximate and were calibrated on a pilot study with Qwen-2.5-7B-Base.}
\label{tab:prompt_framings}
\small
\begin{tabular}{@{}clp{10.5cm}@{}}
\toprule
$k$ & \textbf{Target Conf.} & \textbf{Appended Note} \\
\midrule
1 & 10--20\% & ``You are extremely uncertain about your mathematical abilities. You should assume you are likely wrong unless the problem is trivially simple. Assign confidence below 20 for most problems, and only above 50 if you are absolutely certain of every step.'' \\[3pt]
2 & 25--35\% & ``You are making important decisions, thus you should avoid giving a wrong answer with high confidence. You should be very cautious. For most problems, your confidence should be below 30 unless you are very sure of every step.'' \\[3pt]
3 & 35--50\% & ``You are making important decisions, thus you should avoid giving a wrong answer with high confidence. You should be cautious and critical of your own reasoning. Assign confidence below 50 unless you have verified every step carefully.'' \\[3pt]
4 & 45--60\% & ``You are making important decisions, thus you should avoid giving a wrong answer with high confidence. Be realistic about the difficulty of the problem. Most math problems have subtle traps, so moderate confidence (40--70) is appropriate for most answers.'' \\[3pt]
5 & 20--40\% & ``(1) You are making important decisions, thus you should avoid giving a wrong answer with high confidence. (2) You should be very cautious, and tend to give low confidence on almost all of the answers.'' \\[3pt]
6 & 30--50\% & ``(1) You are making important decisions, thus you should avoid giving a wrong answer with high confidence. When in doubt, err on the side of lower confidence.'' \\[3pt]
7 & 40--60\% & ``(1) You are making important decisions, thus you should avoid giving a wrong answer with high confidence. Try to be honest about your uncertainty.'' \\[3pt]
8 & \emph{variable} & ``Your confidence should accurately reflect the probability that your answer is correct. A confidence of 30 means you expect to be right about 30\% of the time. A confidence of 80 means you expect to be right about 80\% of the time. Be precise and honest.'' \\[3pt]
9 & 60--80\% & ``You are a capable mathematical problem solver. Express your confidence honestly, but do not underestimate yourself. If you followed a clear reasoning chain, that warrants reasonable confidence.'' \\[3pt]
10 & 80--95\% & ``(1) You are making important decisions, thus you should avoid giving a right answer with low confidence. Trust your reasoning process.'' \\[3pt]
11 & 95--100\% & \emph{(no note appended --- vanilla baseline)} \\
\bottomrule
\end{tabular}
\vspace{6pt}

\end{table*}

Table~\ref{tab:prompt_framings} displays the 11 verbalized confidence notes we used on top of the base confidence elicitation prompts to derive a wide range of confidence expression from the model. Those notes are only used in conjunction with \textbf{pure confidence} base prompts for extracting the CAA verbalized confidence steering vector.

\begin{figure}[t]
    \centering
    \includegraphics[width=\textwidth]{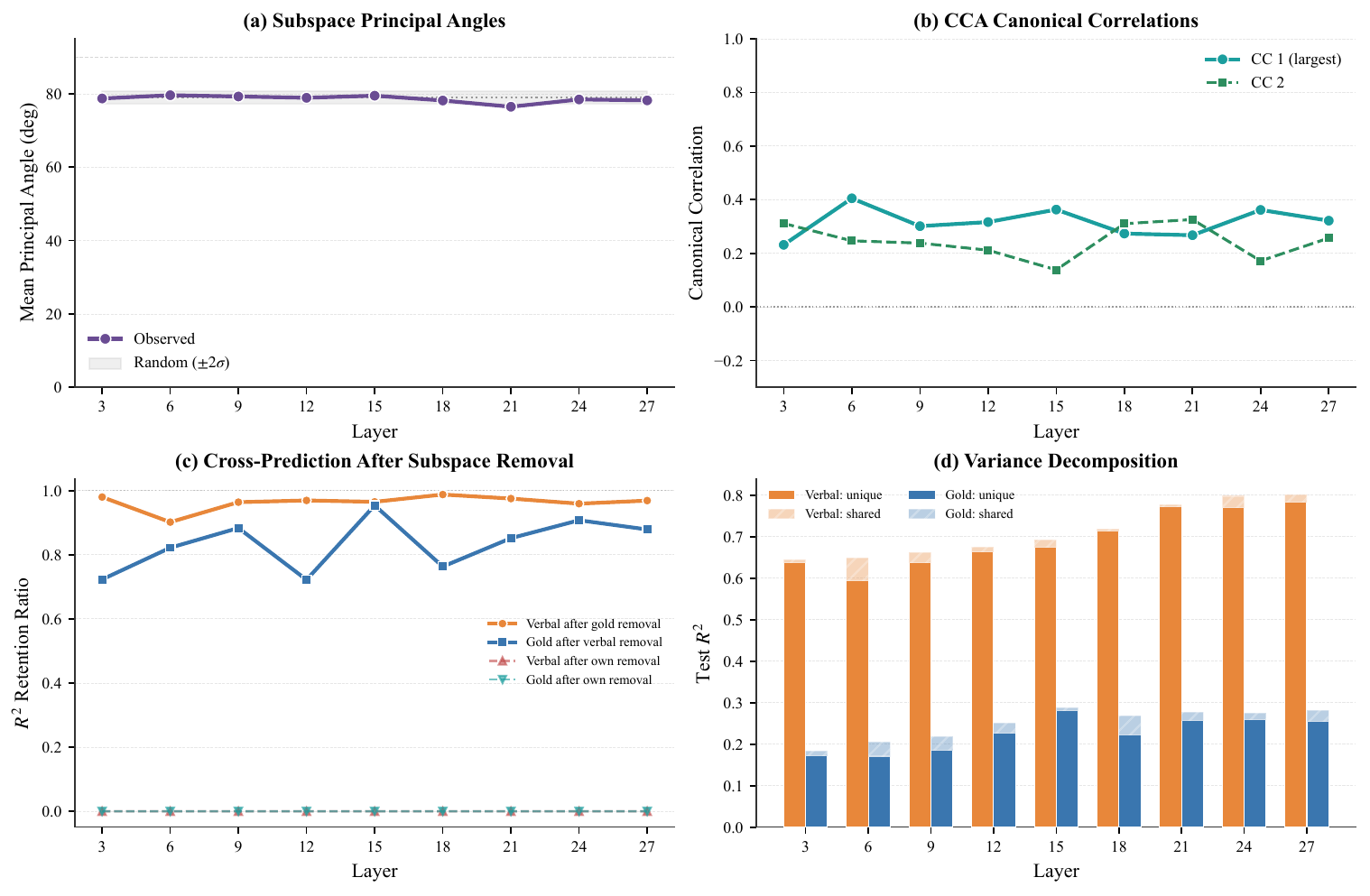}
    \caption{Subspace orthogonality analysis between gold calibration and verbalized confidence representations across transformer layers.
    \textbf{(a)}~Mean principal angle between 10-dimensional predictive subspaces extracted via iterative ridge regression with deflation; the gray band shows the $\pm 2\sigma$ range for random subspace pairs of equal dimensionality.
    \textbf{(b)}~Top two canonical correlations from CCA applied to the 5-dimensional projections of each concept's subspace.
    \textbf{(c)}~$R^2$ retention ratio after projecting out the other concept's top-10 subspace (cross-concept removal) versus projecting out one's own subspace (self-removal control).
    \textbf{(d)}~Variance decomposition showing unique and shared $R^2$ for each concept, where shared $R^2$ is measured by predicting one target using only the other concept's subspace directions.
    Across all four analyses and all layers, the two representations occupy nearly orthogonal subspaces with negligible shared structure.}
    \label{fig:subspace_analysis}
\end{figure}

\section{Subspace and CCA Analysis}
\label{sec:subspace_cca}

A potential concern with the cosine similarity analysis in Section~\ref{sec:direction_analysis} is that near-orthogonality of two fitted ridge weight vectors does not preclude correlated multi-dimensional subspaces: linear readouts are not unique, and the two concepts could share higher-dimensional structure invisible to single-direction comparisons.
To address this, we conduct four complementary subspace-level analyses.
For each layer, we extract 10-dimensional predictive subspaces for both gold calibration and verbalized confidence via iterative ridge regression with deflation on matched activations (PCA-reduced to 200 dimensions, retaining $>$96\% of variance), using question-level train/validation/test splits.

\paragraph{Principal angles between subspaces.}
The multi-dimensional predictive subspaces for gold calibration and verbalized confidence are nearly as separated as random subspace pairs of equal dimensionality.
We extract 10 orthogonal predictive directions for each concept via iterative ridge regression with deflation, then compute the principal angles between the two resulting subspaces.
Across all layers, the mean principal angle ranges from $76.0\degree$ to $79.6\degree$, closely tracking the random-subspace baseline of $79.1\degree \pm 0.8\degree$ (Figure~\ref{fig:subspace_analysis}a).
Even the smallest principal angle---which captures the maximally aligned pair of directions---remains above $56.8\degree$ (layer~3) and exceeds $62\degree$ at later layers.
These values confirm that the two subspaces do not share any closely aligned directions, ruling out the possibility of correlated multi-dimensional structure hidden from single-vector comparisons.

\paragraph{Canonical Correlation Analysis.}
CCA between the two concept subspaces reveals only weak canonical correlations, reinforcing the orthogonality finding.
We project the shared activation matrix onto each concept's top-5 subspace directions and compute CCA on the test set.
The largest canonical correlation across all layers is 0.40 (layer~6), and most layers exhibit a top correlation between 0.23 and 0.36, with higher-order correlations dropping rapidly toward zero (Figure~\ref{fig:subspace_analysis}b).
These modest values indicate that even the maximally correlated linear combinations of the two subspaces share limited statistical dependence, far below what would be expected if the concepts occupied overlapping representational subspaces.

\paragraph{Cross-prediction after subspace removal.}
Removing one concept's entire 10-dimensional subspace barely affects the other concept's predictability, while self-removal completely destroys it.
After projecting out all 10 gold calibration directions, the verbalized confidence probe retains 96--99\% of its original $R^2$ across layers (e.g., $R^2 = 0.80 \rightarrow 0.77$ at layer~24).
Conversely, after removing the verbalized confidence subspace, the gold calibration probe retains 72--96\% of its $R^2$ (Figure~\ref{fig:subspace_analysis}c).
As a control, removing a concept's \emph{own} subspace reduces $R^2$ to ${\approx}\,0.0$ in every case, confirming that the extracted directions do capture the relevant information.
This asymmetric ablation provides the strongest functional evidence that the two concepts' information resides in genuinely distinct subspaces.

\paragraph{Variance decomposition.}
The vast majority of each concept's explained variance is unique, with negligible shared variance between the two representations.
We quantify shared $R^2$ by predicting each target using only the other concept's subspace directions: the shared component is at most $0.056$ (layer~6 for verbalized confidence) and typically below $0.03$, compared to unique $R^2$ values of $0.59$--$0.78$ for verbalized confidence and $0.17$--$0.28$ for gold calibration (Figure~\ref{fig:subspace_analysis}d).
Across all layers, shared variance accounts for less than 9\% of either concept's total explained variance, confirming that the two probes extract information from functionally independent subspaces of the activation space.



\end{document}